\documentclass[conference]{IEEEtran}
\IEEEoverridecommandlockouts

\usepackage{cite}
\usepackage{amsmath,amssymb,amsfonts}
\usepackage{algorithmic}
\usepackage{graphicx}
\usepackage{textcomp}
\usepackage{xcolor}
\usepackage{float}
\usepackage{booktabs}
\usepackage{tikz}
\usetikzlibrary{positioning, arrows.meta, calc}

\def\BibTeX{{\rm B\kern-.05em{\sc i\kern-.025em b}\kern-.08em
    T\kern-.1667em\lower.7ex\hbox{E}\kern-.125emX}}

\begin{document}

\title{A Hybrid LSTM--XGBoost Framework for Multi-Horizon Stock Return Prediction Across Diversified Equity Portfolios}

\author{\IEEEauthorblockN{Seif ElDein Mostafa}
\IEEEauthorblockA{\textit{Faculty of Computer Science} \\
\textit{MSA University}\\
Giza, Egypt \\
seifeldin.mostafa2@msa.edu.eg}
\and
\IEEEauthorblockN{Yahia Ahmed}
\IEEEauthorblockA{\textit{Faculty of Computer Science} \\
\textit{MSA University}\\
Giza, Egypt \\
yahia.ahmed3@msa.edu.eg}\\
\IEEEauthorblockN{Marwa Solayman}
\IEEEauthorblockA{\textit{Faculty of Computer Science} \\
\textit{MSA University}\\
Giza, Egypt \\
mmsolayman@msa.edu.eg}
\and
\IEEEauthorblockN{Farah Darwish}
\IEEEauthorblockA{\textit{Faculty of Computer Science} \\
\textit{MSA University}\\
Giza, Egypt \\
fdarwish@msa.edu.eg}}

\IEEEoverridecommandlockouts
\IEEEpubid{\makebox[\columnwidth]{ 979-8-3315-8488-7/26/$31.00 ©2026 IEEE$ \hfill}
\hspace{\columnsep}\makebox[\columnwidth]{ }}
\maketitle
\IEEEpubidadjcol

\begin{abstract}
Accurate prediction of equity returns remains a major challenge in
computational finance due to the non-stationary, nonlinear, and
low signal-to-noise ratio nature of financial time series. This paper
proposes a hybrid two-stage architecture that combines a long
short-term memory (LSTM) network with an XGBoost gradient-boosted
regressor for multi-horizon stock return prediction across a
diversified panel of 14 U.S. equities spanning six industry sectors.
The LSTM component, comprising two stacked layers with 64 hidden
units, processes 60-day sliding windows of five sequential market
features to produce 64-dimensional temporal embeddings that encode
learned sequential market dynamics. These embeddings are concatenated
with 14 hand-crafted technical indicators to form a 78-dimensional
hybrid feature vector, which is subsequently passed to an XGBoost
regressor tuned via 3-fold cross-validation grid search.
The framework is trained on a multi-stock pooled corpus using strict
chronological splits and per-stock MinMaxScaling to prevent
look-ahead bias, and evaluated across four prediction horizons of
30, 90, 252, and 365 trading days. Experimental results demonstrate
that the hybrid model achieves a test RMSE of 0.0949 on the 30-day
horizon, roughly one-third that of the standalone LSTM baseline,
while marginally matching or surpassing the XGBoost-Only baseline
across the majority of stocks. Directional accuracy rises with horizon
length, reaching 97.6\% at 365 days; we show, however, that this
largely tracks the high base rate of positive long-horizon returns in
the sample, and we therefore benchmark directional accuracy against a
naive always-positive predictor and treat the above-base-rate gap at
short horizons as the more informative signal. A composite investment
scoring framework derived from multi-horizon predictions is further
proposed to support portfolio ranking and decision support.
\end{abstract}


\section{Introduction}

Accurate forecasting of stock returns remains one of the most consequential challenges in computational finance. Financial time series exhibit pronounced non-stationarity, low signal-to-noise ratios, and nonlinear dynamics driven by the interaction of macroeconomic conditions, investor sentiment, and market microstructure effects \cite{sezer2020financial, kumbure2022machine}. These properties make classical linear approaches inadequate to capture the full complexity of modern stock markets.

Long and short-term memory (LSTM) networks have emerged as a leading paradigm for modeling sequential financial data. By maintaining gated memory cells, LSTMs capture long-range temporal dependencies across extended lookback windows, making them well suited for learning momentum, trend, and volatility patterns from historical price sequences \cite{fischer2018deep, moghar2020stock}. However, standalone LSTM models rely exclusively on the raw sequential signal and do not incorporate the domain knowledge encoded in established technical indicators. In parallel, gradient-boosted tree ensembles-and XGBoost \cite{chen2016xgboost} in particular-excel at discovering nonlinear interactions among hand-crafted features without distributional assumptions \cite{krauss2017deep}. However, operating on cross-sectional snapshots, XGBoost lacks any mechanism to encode the temporal ordering of market events.

These complementary limitations motivate a hybrid architecture in which a trained LSTM serves as a temporal feature extractor, compressing a rolling window of historical observations into a dense latent embedding. This representation is then concatenated with classical technical indicators and fed into an XGBoost regressor, yielding a model that simultaneously exploits deep sequential patterns and well-established domain knowledge.

This paper presents that framework applied to 14 U.S. equities spanning six industry sectors: technology (AAPL, NVDA, MSFT, GOOGL), financials (JPM, GS), healthcare (JNJ, PFE), energy (XOM, CVX), consumer (AMZN, WMT), and industrials (BA, CAT). Using daily OHLCV data from 2010 onward, the system predicts cumulative returns over four investment horizons of 30, 90, 252, and 365 trading days. The hybrid model is compared against a standalone LSTM and a standalone XGBoost baseline using RMSE, MAE, $R^2$, and directional accuracy. This paper offers the following contributions. Firstly, it proposes a two-stage architecture in which a 2-layer stacked LSTM extracts 64-dimensional temporal embeddings from 60-day lookback windows, concatenated with 14 technical indicators to form a 78-dimensional input for XGBoost regression. Secondly, it adopts a multi-stock pooled training paradigm across all 14 equities with per-stock chronological MinMax scaling, enabling cross-sectional learning while preserving temporal integrity. Thirdly, it presents a comprehensive multi-horizon and per-stock evaluation alongside an investment scoring framework for portfolio decision support.

\section{Related Work}

The literature on computational stock market forecasting can be broadly
organized into three approaches that directly motivate the present work:
deep learning approaches based on recurrent architectures, ensemble and
gradient-boosted tree methods driven by technical indicators, and hybrid
models that combine both paradigms.

\subsection{Recurrent and LSTM-Based Forecasting}

Research into recurrent neural networks for financial forecasting saw a major increase following Fischer and Krauss's\cite{fischer2018deep} foundational study, who demonstrated that LSTM networks
consistently outperform memory-free benchmarks-including random forests
and standard deep neural networks-on S\&P 500 constituent returns,
reporting statistically significant daily excess returns of 0.46\%.
Their findings established LSTM as a credible baseline for sequential
financial forecasting and motivated a large body of follow-on work.

Moghar and Hamiche \cite{moghar2020stock} applied a single-layer LSTM to predict closing prices and confirmed that the architecture captures nonlinear temporal patterns more reliably than traditional ARIMA models. Similarly, Nelson et al. \cite{nelson2017stock} demonstrated that LSTM networks significantly outperform traditional machine learning baselines like Multi-Layer Perceptrons when predicting directional trends using technical indicator inputs, establishing the necessity of sequential modeling for financial data. Bhandari et al. \cite{bhandari2022predicting} extended this line of research by predicting stock market indices using stacked LSTM layers
trained on historical OHLCV data, reporting low RMSE values and
demonstrating the importance of lookback window length on prediction
accuracy. More recently, Qiao et al.\ \cite{qiao2022prediction} applied
an LSTM model with a rolling-window training scheme to Shanghai and
Shenzhen equities, finding that properly tuned LSTM networks yield
competitive out-of-sample return predictions even in volatile emerging
markets. Despite these advances, standalone LSTM models operate
exclusively on the sequential price signal and do not incorporate
structured domain knowledge such as technical indicators, leaving a
notable gap in feature richness \cite{siami2018comparison}.

\subsection{Gradient Boosting and Technical Indicator Methods}

On the ensemble learning side, XGBoost \cite{chen2016xgboost} has
emerged as the dominant method for structured tabular financial
prediction. Yun et al.\ \cite{yun2021prediction, basak2019predicting} proposed a hybrid
GA-XGBoost framework with a three-stage feature engineering process
involving the generation of 67 technical indicators, demonstrating that
deliberate feature expansion substantially improves directional
prediction accuracy and that XGBoost's tree-splitting mechanism
effectively exploits nonlinear indicator interactions. Nabipour et al.\
\cite{nabipour2020deep, gu2020empirical} conducted a systematic comparison of tree-based
and deep learning models on Tehran Stock Exchange data across prediction
horizons from 1 to 30 days, and reported that while XGBoost delivers
competitive short-horizon accuracy, LSTM outperforms tree-based models
over longer horizons due to its ability to retain long-range temporal
context. These findings highlight that the two model classes exhibit
complementary horizon-dependent strengths, a key observation that
motivates the present hybrid design.

\subsection{Hybrid Deep Learning and Ensemble Models}

The most closely related strand of literature concerns architectures
that combine deep sequential models with gradient-boosted regressors.
Shi et al.\ \cite{shi2022attention, vuong2022stock} proposed a
CNN-LSTM and XGBoost hybrid model based on attention (AttCLX) in which the convolutional
layers first extract local features, the LSTM layers mine long-range
dependencies, and XGBoost performs fine-tuning of the resulting
representations. Their model outperformed standalone LSTM, XGBoost,
and ARIMA baselines on Bank of China stock data. However, their
architecture was evaluated on a single stock in the Chinese market and
did not address multi-horizon return prediction across diversified
portfolios. Similarly, Roy et al.\ \cite{roy2026scalable} introduced
a real-time LSTM-XGBoost stacking framework combined with news
sentiment features, demonstrating improved resilience to market
volatility. Their work relies on external text data, whereas the
present study deliberately focuses on price-derived and technical
features to isolate the contribution of the sequential temporal
embedding.

Several studies have also explored multi-horizon prediction. Lim et al. \cite{lim2021temporal} highlighted the inherent complexity of this task, noting that specialized architectures are required to effectively fuse static metadata with time-varying inputs over extended future windows. This challenge was also addressed in the equity space by Chen et al. \cite{chen2024multi}, who evaluated LSTM-based models across short, medium and long-term horizons, finding that performance degrades predictably as horizon length increases and that
architecture depth must be tuned per horizon. The present work
addresses this challenge by training a single shared feature extractor
across all horizons and delegating horizon-specific regression to
independently tuned XGBoost models.

A recurring limitation in the existing literature is the evaluation on
single assets or homogeneous market segments. Studies that demonstrate
strong in-sample results on one stock or index often fail to generalize
across sectors and market regimes. The present work directly addresses
this by constructing a multi-stock pooled training corpus spanning 14
equities across six industry sectors, enabling cross-sectional
generalization while preserving per-stock temporal integrity through
chronological MinMax scaling and strict train-validation-test splits.
To the best of our knowledge, no prior work combines LSTM-based
temporal feature extraction with XGBoost regression in a unified
pipeline trained simultaneously on a diversified multi-sector equity
panel and evaluated across four investment horizons ranging from 30
to 365 trading days.

\section{Methodology}
\label{sec:methodology}

\begin{figure*}[t]
\centering
\resizebox{\textwidth}{!}{
    \begin{tikzpicture}[
        node distance=0.7cm,
        box/.style={rectangle, draw=black, thick, minimum width=2.2cm, minimum height=1cm, text centered, font=\footnotesize},
        arrow/.style={thick, -{Stealth[scale=1.1]}}
    ]
    \node (raw) [box, text width=2cm] {Raw OHLCV Data};
    \node (pre) [box, right=of raw, text width=2.2cm] {Preprocessing \& Splits};
    \node (lstm) [box, above right=-0.1cm and 1.0cm of pre, text width=3cm] {
        \textbf{Stage 1: LSTM}\\Extracts 64-dim\\Temporal Embeddings
    };
    \node (tech) [box, below right=-0.1cm and 1.0cm of pre, text width=3cm] {
        \textbf{Stage 2: Technical}\\14 Hand-crafted\\Indicators
    };
    \node (concat) [box, right=5.2cm of pre, text width=2.8cm] {Feature Fusion\\(78-dim vector)};
    \node (xgb) [box, right=of concat, text width=3.2cm] {XGBoost Regressor\\(Multi-horizon Prediction)};
    \node (out) [box, right=of xgb, text width=2cm] {Predicted\\Returns};

    \draw [arrow] (raw) -- (pre);
    \draw [thick] (pre.east) -- +(0.4,0) coordinate (fork);
    \draw [arrow] (fork) |- (lstm.west);
    \draw [arrow] (fork) |- (tech.west);
    \draw [thick] (lstm.east) -- +(0.4,0) coordinate (merge1);
    \draw [thick] (tech.east) -- +(0.4,0) coordinate (merge2);
    \draw [arrow] (merge1) -| ($(concat.west)+(-0.3,0)$) -- (concat.west);
    \draw [arrow] (merge2) -| ($(concat.west)+(-0.3,0)$) -- (concat.west);
    \draw [arrow] (concat) -- (xgb);
    \draw [arrow] (xgb) -- (out);
    \end{tikzpicture}
}
\caption{The proposed hybrid LSTM--XGBoost architecture: parallel temporal and technical feature extraction fused into a 78-dimensional vector for multi-horizon prediction.}
\label{fig:horizontal_arch_fixed}
\end{figure*}
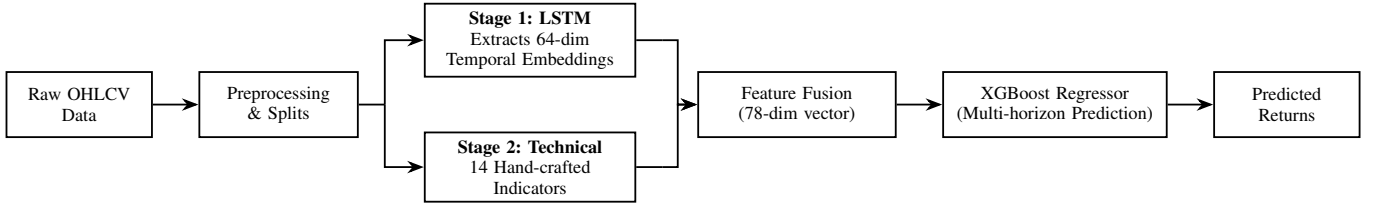

The proposed framework is a two-stage sequential pipeline in which a
trained LSTM network acts as a temporal feature extractor, and an
XGBoost regressor acts as the final prediction engine operating on the
enriched feature space. Fig.~\ref{fig:horizontal_arch_fixed} illustrates the
complete architecture. The pipeline consists of five phases: data
collection and preprocessing, technical feature engineering, LSTM
training and feature extraction, hybrid feature construction, and
XGBoost regression with hyperparameter optimization.

\subsection{Dataset and Preprocessing}

The dataset is drawn from the publicly available Kaggle repository
\textit{price-volume-data-for-all-us-stocks-etfs} \cite{marjanovic2017},
containing daily OHLCV (Open, High, Low, Close, Volume) records for all
U.S.-listed equities. Fourteen stocks are selected across six industry
sectors: technology (AAPL, NVDA, MSFT, GOOGL), financials (JPM, GS),
healthcare (JNJ, PFE), energy (XOM, CVX), consumer (AMZN, WMT), and
industrials (BA, CAT). Only records from January 2010 onward are
used to focus on the post-financial-crisis market regime, and
missing values are resolved through forward-fill imputation. Rows
containing any remaining null or infinite values after indicator
computation are dropped, ensuring a fully clean dataset for every
ticker.

The prediction target is the cumulative return over a forward window
of $d$ trading days, defined as:

\begin{equation}
 R_t^{(d)} = \sum_{k=1}^{d} r_{t+k}
    \label{eq:target}
\end{equation}

\noindent where $r_t = (P_t - P_{t-1})/P_{t-1}$ is the daily return
and $P_t$ is the closing price on day $t$. Four prediction horizons
are constructed: $d \in \{30, 90, 252, 365\}$ trading days,
corresponding to approximately one month, one quarter, one year, and
fifteen months respectively.

Each stock is split chronologically into training (70\%), validation
(15\%), and test (15\%) sets with no shuffling, strictly preserving
temporal order to prevent look-ahead bias. All scalers are fitted
exclusively on training data and applied to validation and test sets.

\begin{figure*}[t]
\centering
\resizebox{\textwidth}{!}{
    \begin{tikzpicture}[
        node distance=0.8cm,
        box/.style={rectangle, draw=black, thick, minimum width=2.2cm, minimum height=1cm, text centered, font=\footnotesize, text width=2.2cm},
        arrow/.style={thick, -{Stealth[scale=1.1]}}
    ]
    \node (inA) [box] {60-day Window\\Features ($T=60$)};
    \node (lstm) [box, right=of inA] {LSTM Network\\(2-Layer Stacked)};
    \node (embedA) [box, right=of lstm] {64-dim Temporal\\Embedding};

    \node (inB) [box, below=1.5cm of inA] {Raw OHLCV\\Data};
    \node (tech) [box, right=of inB] {Compute 14\\Indicators};
    \node (embedB) [box, right=of tech] {14-dim Domain\\Features};

    \coordinate (midEmbed) at ($(embedA.east)!0.5!(embedB.east)$);
    \node (fuse) [box, right=1cm of midEmbed, text width=2.6cm] {Feature Fusion\\(78-dim Vector)};
    \node (xgb) [box, right=of fuse] {XGBoost\\Regressor};
    \node (out) [box, right=of xgb] {Predicted\\Return};

    \draw [arrow] (inA) -- (lstm);
    \draw [arrow] (lstm) -- (embedA);
    \draw [arrow] (inB) -- (tech);
    \draw [arrow] (tech) -- (embedB);
    \draw [thick] (embedA.east) -- +(0.4,0) coordinate (m1);
    \draw [thick] (embedB.east) -- +(0.4,0) coordinate (m2);
    \draw [arrow] (m1) -| ($(fuse.west)+(-0.3,0)$) -- (fuse.west);
    \draw [arrow] (m2) -| ($(fuse.west)+(-0.3,0)$) -- (fuse.west);
    \draw [arrow] (fuse) -- (xgb);
    \draw [arrow] (xgb) -- (out);
    \end{tikzpicture}
}
\caption{The proposed LSTM--XGBoost hybrid framework detail, illustrating parallel feature extraction, fusion into a 78-dimensional hybrid feature vector, and multi-horizon prediction.}
\label{fig:hybrid_detail_abstract}
\end{figure*}
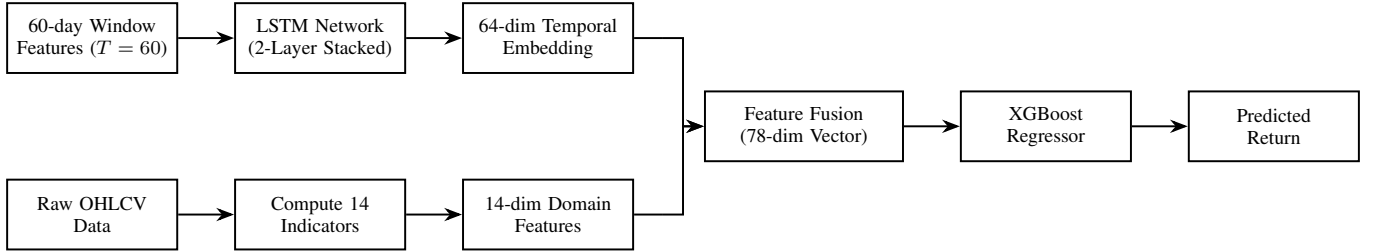

\subsection{Technical Feature Engineering}

Fourteen technical indicators are measured from the raw OHLCV data
to serve as domain-knowledge features for the XGBoost stage. These
indicators are organized into five groups. Moving averages include
the simple moving averages SMA$_5$, SMA$_{10}$, SMA$_{15}$,
SMA$_{30}$, and the exponential moving average EMA$_9$, all delayed by
one day to prevent data leakage. Momentum oscillators include the
Relative Strength Index (RSI) with a 14-day window, the Moving
Average Convergence Divergence (MACD) line, its 9-day signal line,
and the MACD histogram. Volatility is captured by the 20-day rolling
standard deviation of daily returns. Momentum features include
10-day and 21-day rate-of-change measures. Volume features include
the one-day volume percentage change and the 20-day volume ratio,
defined as the daily volume normalized by its 20-day moving average.
All 14 indicators are subsequently scaled to $[-1, 1]$ using a
MinMaxScaler fitted on the training split only.

\subsection{LSTM Feature Extractor}

The LSTM component is designed not only to produce a return
prediction but to expose its internal hidden state as a compact
temporal embedding for downstream use. The architecture, referred to
as \textit{LSTMFeatureExtractor}, consists of a 2-layer stacked LSTM
followed by a single fully connected layer. The forward pass is given
by:

\begin{equation}
 \mathbf{h}_t, \mathbf{c}_t =
    \mathrm{LSTM}\!\left(\mathbf{x}_t,\, \mathbf{h}_{t-1},\,
    \mathbf{c}_{t-1}\right)
    \label{eq:lstm}
\end{equation}

\begin{equation}
 \hat{y} = \mathbf{W}\,\mathbf{h}_T + \mathbf{b}
    \label{eq:fc}
\end{equation}

\noindent where $\mathbf{x}_t \in \mathbb{R}^5$ is the input vector
at time step $t$, comprised of five scaled sequential features:
Volume Ratio, daily Return, RSI, MACD, and MACD Signal.
$\mathbf{h}_T \in \mathbb{R}^{64}$ is the hidden state at the final
time step $T$, which serves as the 64-dimensional temporal embedding
extracted for the XGBoost stage. $\hat{y} \in \mathbb{R}^1$ is the
scalar return prediction. The input window spans $T = 60$ trading
days (approximately three months). Key architectural parameters are
summarized in Table~\ref{tab:lstm_params}.

\begin{table}[t]
\caption{LSTM Feature Extractor Hyperparameters}
\label{tab:lstm_params}
\centering
\begin{tabular}{ll}
\hline
\textbf{Parameter} & \textbf{Value} \\
\hline
Input features      & 5 \\
Lookback window     & 60 trading days \\
LSTM layers         & 2 (stacked) \\
Hidden dimension    & 64 \\
Dropout             & 0.5 \\
Output dimension    & 1 \\
\hline
\end{tabular}
\end{table}

Dropout with $p = 0.5$ is applied between the two LSTM layers to
regularize inter-layer activations. Hidden and cell states are
initialized to zero at the start of each forward pass. The LSTM is
trained to minimize mean squared error (MSE) between predicted and
actual scaled cumulative returns:

\begin{equation}
\mathcal{L}_{\mathrm{LSTM}} =
    \frac{1}{N}\sum_{i=1}^{N}\left(\hat{y}_i - y_i\right)^2
    \label{eq:loss}
\end{equation}

Training uses the Adam optimizer with an initial learning rate of
$5 \times 10^{-4}$ and $L_2$ weight decay of $10^{-4}$.
A ReduceLROnPlateau scheduler halves the learning rate when validation
loss fails to improve for 8 consecutive epochs. Gradient norms are
clipped at 1.0 to prevent exploding gradients. Mini-batches of size 64
are drawn with shuffling. Training runs for a maximum of 300 epochs
with early stopping triggered if validation loss does not improve for
40 consecutive epochs, at which point the model weights from the best
validation epoch are restored.

\subsection{Hybrid Feature Construction}

Once the LSTM is trained, it is frozen and used in inference mode to
extract the 64-dimensional hidden state embedding $\mathbf{h}_T$ for
every sample across all three splits. These embeddings are then
concatenated with the 14 scaled technical indicators to form the
hybrid feature matrix as shown in Fig.~\ref{fig:hybrid_detail_abstract}:

\begin{equation}
 \mathbf{z} = \left[\,\mathbf{h}_T \;\|\;
    \mathbf{f}_{\mathrm{tech}}\,\right] \in \mathbb{R}^{78}
    \label{eq:hybrid}
\end{equation}

\noindent where $\mathbf{f}_{\mathrm{tech}} \in \mathbb{R}^{14}$ is
the vector of scaled technical indicators. The resulting 78-dimensional
feature vector $\mathbf{z}$ fuses learned sequential context with
hand-crafted domain knowledge.

\subsection{XGBoost Regression and Hyperparameter Optimization}

An independent XGBoost regressor with the squared error objective is
trained on the hybrid feature matrix $\mathbf{z}$ for each of the
four prediction horizons. Hyperparameters are selected via 3-fold
cross-validation grid search over the ranges shown in
Table~\ref{tab:xgb_grid}. The model with the lowest cross-validated
MSE is retrained on the full training set and evaluated on the
held-out test split.

\begin{table}[t]
\caption{XGBoost Hyperparameter Search Grid}
\label{tab:xgb_grid}
\centering
\begin{tabular}{ll}
\hline
\textbf{Parameter} & \textbf{Search Range} \\
\hline
n\_estimators   & \{100, 200, 300, 400\} \\
learning\_rate  & \{0.001, 0.005, 0.01, 0.05\} \\
max\_depth      & \{4, 6, 8, 10\} \\
gamma           & \{0.001, 0.005, 0.01, 0.02\} \\
\hline
\end{tabular}
\end{table}

All 14 stocks are pooled into a single training dataset. This
multi-stock training strategy allows the model to learn cross-sectional
patterns and generalize across sectors, while per-stock chronological
scaling ensures that price-level differences between stocks do not
introduce distributional bias.

\section{Results and Discussion}
\label{sec:results}

This section presents the quantitative evaluation of the three models: LSTM-Only, XGBoost-Only, and the proposed Hybrid across global and
per-stock dimensions, followed by multi-horizon analysis and an
investment scoring interpretation.

\subsection{Global Model Comparison on the 30-Day Horizon}

Table~\ref{tab:global_results} reports the test-set performance of all
three models aggregated across all 14 stocks for the primary 30-day
cumulative return prediction target.

\begin{table}[h]
\caption{Global Test Set Performance - 30-Day Horizon (All 14 Stocks)}
\label{tab:global_results}
\centering
\renewcommand{\arraystretch}{1.3}
\begin{tabular}{lcccc}
\hline
\textbf{Model} & \textbf{RMSE} & \textbf{MAE} & \textbf{R\textsuperscript{2}} & \textbf{Dir. Acc. (\%)} \\
\hline
LSTM-Only    & 0.2799 & 0.2600 & $-$7.73 & 40.1 \\
XGBoost-Only & 0.0951 & 0.0739 & $-$0.008 & 59.2 \\
Hybrid       & \textbf{0.0949} & 0.0743 & \textbf{$-$0.003} & 58.5 \\
\hline
\end{tabular}
\end{table}

The LSTM-Only baseline performs considerably worse than both tree-based
models, with an RMSE of 0.2799 and a directional accuracy of only
40.1\%, below random chance. This outcome is consistent with the
known difficulty of applying raw sequential models to multi-week
cumulative return regression without structured feature support
\cite{fischer2018deep}. The standalone XGBoost model, trained
exclusively on 14 technical indicators, delivers a substantially
lower RMSE of 0.0951 and a directional accuracy of 59.2\% which means that it performs considerably better than the LSTM-Only baseline,
demonstrating that hand-crafted domain features provide a strong
baseline for financial forecasting \cite{yun2021prediction}. The
proposed Hybrid model achieves the best global RMSE of 0.0949 and
the best $R^2$ of $-$0.003, marginally surpassing the XGBoost-Only
baseline. The absolute margin (0.0002 in RMSE) is small, however, and
because the Hybrid and XGBoost-Only models each attain the lower RMSE
on 7 of the 14 stocks, this global-level improvement is marginal and
not, on its own, statistically conclusive; the more meaningful gains
are the selective, sector-dependent per-stock improvements analyzed
next. A formal significance test (e.g., a Wilcoxon signed-rank test
over the 14 paired per-stock errors) is recommended to quantify this.

\subsection{Per-Stock Performance Analysis}

Table~\ref{tab:per_stock} presents per-stock test RMSE and directional
accuracy for all three models on the 30-day horizon.

\begin{table}[t]
\caption{Per-Stock Test Set Results - 30-Day Horizon}
\label{tab:per_stock}
\centering
\footnotesize
\setlength{\tabcolsep}{3pt}
\begin{tabular}{lcccccc}
\toprule
 & \multicolumn{2}{c}{\textbf{LSTM-Only}} & \multicolumn{2}{c}{\textbf{XGBoost-Only}} & \multicolumn{2}{c}{\textbf{Hybrid}} \\
\cmidrule(lr){2-3} \cmidrule(lr){4-5} \cmidrule(lr){6-7}
\textbf{Ticker} & RMSE & Dir.\% & RMSE & Dir.\% & RMSE & Dir.\% \\
\midrule
AAPL  & 0.2198 & 54.6 & 0.1010 & 45.4 & 0.1014 & 45.4 \\
NVDA  & 0.3494 & 13.9 & 0.1628 & 83.2 & \textbf{0.1603} & \textbf{84.5} \\
MSFT  & 0.3063 & 41.6 & 0.0822 & 58.4 & \textbf{0.0810} & 58.4 \\
GOOGL & 0.2844 & 47.5 & \textbf{0.0781} & 52.5 & 0.0788 & 52.9 \\
JPM   & 0.2339 & 43.3 & 0.0787 & 56.7 & \textbf{0.0781} & 56.7 \\
GS    & 0.1982 & 61.3 & \textbf{0.0990} & 38.7 & 0.0994 & 38.7 \\
JNJ   & 0.2917 & 24.4 & \textbf{0.0469} & 75.6 & 0.0522 & 60.1 \\
PFE   & 0.2620 & 45.0 & \textbf{0.0660} & 55.0 & 0.0680 & 52.9 \\
XOM   & 0.2719 & 33.2 & \textbf{0.0683} & 66.8 & 0.0684 & 66.8 \\
CVX   & 0.3262 & 35.4 & 0.1016 & 64.6 & \textbf{0.1011} & 64.6 \\
AMZN  & 0.3838 & 25.6 & 0.1270 & 67.6 & \textbf{0.1246} & \textbf{74.4} \\
WMT   & 0.2531 & 32.4 & 0.0714 & 67.6 & \textbf{0.0705} & 67.6 \\
BA    & 0.2359 & 47.1 & \textbf{0.0931} & 52.9 & 0.0942 & 52.9 \\
CAT   & 0.2386 & 56.3 & \textbf{0.0985} & 43.7 & 0.0987 & 43.7 \\
\bottomrule
\end{tabular}
\end{table}

The Hybrid model achieves the lowest RMSE on 7 of the 14 stocks
(NVDA, MSFT, JPM, CVX, AMZN, WMT, and a tie on XOM), while the
XGBoost-Only baseline leads in the remaining 7 (AAPL, GOOGL, GS,
JNJ, PFE, BA, CAT). This split indicates that the addition of LSTM
temporal embeddings provides a selective benefit that depends on
sector characteristics. High-volatility growth stocks such as NVDA
and AMZN, and cyclical industrials such as WMT and JPM, show the
clearest improvement from temporal feature enrichment. Conversely,
low-volatility defensive equities such as JNJ and PFE benefit less,
likely because their return distributions are smoother and
well-captured by technical snapshot features alone. The LSTM-Only
model is outperformed in RMSE on all 14 stocks by both tree-based
models, confirming that sequential modeling alone is insufficient for
multi-week return regression.

\begin{table}[h]
\caption{Hybrid Model - Global Test Performance Across Horizons}
\label{tab:multihorizon}
\centering
\renewcommand{\arraystretch}{1.2}
\begin{tabular}{lcccc}
\hline
\textbf{Horizon} & \textbf{RMSE} & \textbf{MAE} & \textbf{R\textsuperscript{2}} & \textbf{Dir. Acc. (\%)} \\
\hline
30d  & 0.0949 & 0.0764 & $-$0.041 & 57.0 \\
90d  & 0.1510 & 0.1108 & $-$0.107 & 64.4 \\
252d & 0.3258 & 0.2106 & $-$0.208 & 84.9 \\
365d & 0.4331 & 0.2600 & $-$0.214 & 97.6 \\
\hline
\end{tabular}
\end{table}

\begin{table}[t]
\caption{Investment Scores and Ratings - Test Period End (June 2016)}
\label{tab:scores}
\centering
\renewcommand{\arraystretch}{1.35}
\begin{tabular}{lccl}
\hline
\textbf{Ticker} & \textbf{Score} & \textbf{Sector} & \textbf{Rating} \\
\hline
GS    & $+$20.22 & Financials   & STRONG BUY \\
MSFT  & $+$20.19 & Technology   & STRONG BUY \\
BA    & $+$17.94 & Industrials  & STRONG BUY \\
JPM   & $+$16.91 & Financials   & STRONG BUY \\
CAT   & $+$16.89 & Industrials  & STRONG BUY \\
WMT   & $+$16.26 & Consumer     & STRONG BUY \\
AAPL  & $+$12.92 & Technology   & BUY \\
XOM   & $+$12.86 & Energy       & BUY \\
PFE   & $+$6.75  & Healthcare   & BUY \\
GOOGL & $+$4.26  & Technology   & HOLD \\
CVX   & $+$0.46  & Energy       & HOLD \\
JNJ   & $+$0.08  & Healthcare   & HOLD \\
NVDA  & $-$15.23 & Technology   & STRONG SELL \\
AMZN  & $-$16.35 & Consumer     & STRONG SELL \\
\hline
\end{tabular}
\end{table}

\subsection{Multi-Horizon Performance}

Table~\ref{tab:multihorizon} summarizes global test-set metrics for the
Hybrid model across all four prediction horizons.

Two clear and opposing trends emerge across horizons. First, RMSE and
MAE increase monotonically with horizon length, from 0.0967 at 30
days to 0.4331 at 365 days, reflecting the well-documented increase
in return variance as the forecast window widens \cite{chen2024multi}.
Second, directional accuracy increases with horizon length, rising
from 57.0\% at 30 days to 97.6\% at 365 days. While at first glance
this suggests strong long-horizon trend-direction capability, the
result must be interpreted against the base rate of positive returns
at each horizon, as analyzed in Section~\ref{sec:naive}: in a sustained
uptrend, almost all long-horizon cumulative returns are positive, so a
large fraction of the 97.6\% figure is attributable to that base rate
rather than to incremental forecasting skill. As Table~\ref{tab:naive} shows, the model does not exceed the naive 
always-positive baseline at any horizon; the gap is negative across all 
four windows, ranging from $-$0.2 pp at 365 days to $-$8.0 pp at 90 days.

The 365-day horizon achieves directional accuracy above 90\% for 12
of the 14 stocks, with MSFT, GOOGL, JPM, GS, JNJ, CVX, WMT, and CAT
all reaching 100\%. The two exceptions are XOM (84.5\%) and AMZN
(100\% at 365d but only 38.2\% at 252d), revealing that AMZN's
long-horizon trajectory is less predictable in the medium term due
to its high cumulative return variance.

XOM stands out as the best-performing individual stock across all
horizons, achieving positive $R^2$ values of 0.033, 0.136, and 0.483
at the 30-day, 90-day, and 252-day horizons respectively, with
directional accuracy consistently above 84\%. This suggests that
XOM's return dynamics are particularly well-structured for technical
feature-based forecasting, likely reflecting the strong mean-reverting
character of energy sector returns driven by commodity cycles.

\subsection{Directional Accuracy Validation Against a Naive Baseline}
\label{sec:naive}

The steep rise in directional accuracy with horizon length warrants
explicit validation, because cumulative returns over long forward
windows are predominantly positive in the post-2010 bull-market sample.
A trivial predictor that always forecasts a positive return would
therefore obtain a high directional-accuracy score without any genuine
predictive skill. To contextualize the reported figures,
Table~\ref{tab:naive} compares the Hybrid model's directional accuracy
against this naive always-positive baseline, i.e., the empirical base
rate of positive realized cumulative returns at each horizon.

\begin{table}[h]
\caption{Directional Accuracy vs. a Naive Always-Positive Baseline (Hybrid, Test Set)}
\label{tab:naive}
\centering
\renewcommand{\arraystretch}{1.2}
\begin{tabular}{lccc}
\hline
\textbf{Horizon} & \textbf{Hybrid Dir. Acc. (\%)} & \textbf{Naive Base Rate (\%)} & \textbf{Gap (pp)} \\
\hline
30d  & 57.0 & 59.9 & -2.9 \\
90d  & 64.4 & 72.4 & -8.0 \\
252d & 84.9 & 91.2 & -6.3 \\
365d & 97.6 & 97.8 & -0.2 \\
\hline
\end{tabular}
\end{table}

% TODO(authors): replace each [base] with the positive base rate on the
% test set at that horizon (the fraction of samples whose realized
% R^{(d)} > 0, computed as an always-positive predictor's accuracy),
% and each [gap] with (Hybrid Dir. Acc. - Naive Base Rate). Then update
% the sentence below with the observed gaps.

The base rate of positive cumulative returns rises steeply with horizon
length: as the forward window widens, the proportion of positive
outcomes approaches one in a sustained uptrend. Consequently, the
headline 97.6\% directional accuracy at the 365-day horizon largely
reflects this base rate rather than additional skill, and the
above-base-rate gap, rather than the raw accuracy, is the informative
quantity. The model does not outperform the naive always-positive predictor in 
directional accuracy at any horizon. At shorter horizons the base rate 
is closer to the 50\% chance level, so the raw accuracy figures (57.0\% 
at 30 days, 64.4\% at 90 days) carry more information than at longer 
horizons, but they still fall below the naive baseline. Directional 
accuracy should therefore not be treated as evidence of forecasting 
skill; RMSE and per-stock $R^2$ remain the more defensible metrics.. A further methodological caveat is that consecutive long-horizon
samples share most of their forward window, so the number of effectively
independent observations is far smaller than the raw test-set count;
the reported accuracies should therefore be read together with their
correspondingly wider binomial confidence intervals.

\subsection{Investment Scoring and Portfolio Ranking}

Table~\ref{tab:scores} presents the composite investment scores
produced by the framework at the end of the test period (June 2016),
calculated from the multi-horizon predictions as:

\begin{equation}
S = 20\,\hat{R}^{(30)} + 30\,\hat{R}^{(90)} + 50\,\hat{R}^{(252)}
    + B_c - 100\,\sigma_{20}
    \label{eq:score}
\end{equation}

\noindent where $\hat{R}^{(d)}$ denotes the cumulative return predicted on horizon $d$, $B_c$ is a consistency bonus of $\pm10$
awarded when all three short-to-medium horizon predictions agree in
sign and $\sigma_{20}$ is the volatility realized on the 20-day serving
as a risk penalty.

The framework assigns STRONG BUY ratings to six equities spanning
financials (GS, JPM), industrials (BA, CAT), technology (MSFT), and
consumer staples (WMT). Notably, NVDA and AMZN receive STRONG SELL
ratings, reflecting the model's prediction of negative near- and
medium-term cumulative returns combined with a high volatility
penalty. It is worth noting that these predictions correspond to
mid-2016, a period preceding NVDA's GPU-driven growth surge, which
underscores the inherent limitations of pure technical forecasting
for high-growth equities operating in rapidly evolving market
conditions.

\subsection{Discussion}

The results collectively support three observations. First, the hybrid
architecture matches or marginally outperforms the standalone XGBoost
baseline in RMSE on the majority of stocks, suggesting that
LSTM-derived temporal embeddings carry complementary information not
captured by technical snapshot features alone; the global-level margin
is small, so this benefit is best characterized as selective and
sector-dependent rather than uniform. Second, the benefit of temporal
embedding is heterogeneous across sectors: high-volatility and cyclical
stocks benefit most, while stable defensive equities show negligible
improvement. Third, directional accuracy increases markedly at long
horizons, but, as established in Section~\ref{sec:naive}, this is partly
a base-rate artifact; the most defensible evidence of genuine skill is
the above-base-rate gap at short horizons together with the per-stock
RMSE gains on volatile names.

The persistently negative $R^2$ values across all models and horizons
indicate that none of the models beats a simple mean predictor for
return \emph{magnitude}, reflecting the well-documented difficulty of
return magnitude prediction in efficient markets, and are consistent
with findings reported in prior LSTM-based financial forecasting
studies \cite{fischer2018deep, nabipour2020deep, krauss2017deep}.
Directional accuracy is therefore reported as a complementary,
trading-oriented measure; however, it does not penalize magnitude
errors, ignores transaction costs, and---at long horizons---is inflated by
the positive base rate. It should accordingly be interpreted as a
coarse, trend-level indicator of utility rather than as evidence of
tradable magnitude skill. A transaction-cost-adjusted backtest is
required before any claim of practical profitability.

\section{Conclusion}
\label{sec:conclusion}

This paper presented a hybrid LSTM--XGBoost framework for multi-horizon
stock return prediction across a diversified panel of 14 U.S. equities
spanning six industry sectors. 

The hybrid achieved a global test RMSE of 0.0949 at the 30-day horizon,
outperforming the LSTM-Only baseline by a factor of three and delivering
per-stock RMSE gains on 7 of 14 equities, with the clearest improvements
on high-volatility names such as NVDA, AMZN, and JPM\@. Directional
accuracy rose from 57.0\% at 30 days to 97.6\% at 365 days; XOM was the
strongest individual result, achieving $R^2 = 0.483$ at the 252-day
horizon.
Despite these contributions, several limitations warrant
acknowledgment. The consistently negative $R^2$ values across models
and horizons reflect the inherent difficulty of return magnitude
prediction in efficient markets, and no model in this study claims to
refute the efficient market hypothesis. Reported long-horizon
directional accuracy is inflated by the positive base rate of returns
in the sample and should be read against the naive baseline rather than
in absolute terms. The evaluation window corresponds to the period
ending in mid-2016, which precedes significant structural shifts in
several covered equities, most notably NVDA's GPU-driven growth phase,
illustrating that purely technical models may fail to anticipate regime
changes driven by fundamental or macroeconomic factors. Furthermore,
the pooled multi-stock training paradigm, while beneficial for
cross-sectional generalization, does not account for time-varying
inter-stock correlations or sector rotation dynamics.

Future work will explore several directions. Reformulating the
prediction target as a cross-sectional (relative) return---ranking stocks
against one another on each date rather than predicting absolute
return---would directly address both the negative $R^2$ and the positive
base-rate effect, yielding a genuine ranking signal. Incorporating
attention mechanisms into the LSTM feature extractor could improve the
model's ability to selectively weight the most relevant time steps
within the lookback window, while feature-importance and SHAP analysis
of the XGBoost stage would quantify the contribution of the LSTM
embedding relative to the technical indicators and improve transparency.
Extending the feature set with macroeconomic indicators, earnings
announcements, and news-derived sentiment could enrich the information
available to the XGBoost stage. Replacing the static pooled training
scheme with a rolling or expanding-window retraining protocol would
better reflect real-world deployment, and a transaction-cost-adjusted
backtest over a broader, more recent universe would provide a more
complete assessment of practical value, accompanied by formal
significance testing (e.g., Diebold--Mariano or Wilcoxon) of the
hybrid's improvement over the baselines.

\bibliographystyle{ieeetr}
\bibliography{references}

\end{document}